\documentclass[10pt,twocolumn,letterpaper]{article}

 \usepackage[pagenumbers]{cvpr} 

\usepackage[dvipsnames]{xcolor}

\usepackage{amsmath}
\usepackage{amssymb}
\usepackage{nicematrix}
\newcommand{\RNum}[1]{\lowercase\expandafter{\romannumeral #1\relax}}

\usepackage{booktabs}
\usepackage{tabularx}
\usepackage{adjustbox}

\definecolor{cvprblue}{rgb}{0.21,0.49,0.74}
\usepackage[pagebackref,breaklinks,colorlinks,citecolor=cvprblue]{hyperref}

\def\paperID{12908} 
\def\confName{CVPR}
\def\confYear{2024}

\title{Exploring the Relationship between Samples and Masks for Robust Defect Localization}

\iftrue
\author{Jiang Lin\\
PALM Lab, Southeast University\\
{\tt\small 220215663@seu.edu.cn}
\and
Yaping Yan\\
PALM Lab, Southeast University\\
{\tt\small yan@seu.edu.cn}
}
\fi

\begin{document}
\maketitle

\begin{abstract}
Defect detection aims to detect and localize regions out of the normal distribution.
Previous approaches model normality and compare it with the input to identify defective regions, potentially limiting their generalizability.
This paper proposes a one-stage framework that detects defective patterns directly without the modeling process.
This ability is adopted through the joint efforts of three parties: a generative adversarial network (GAN), a newly proposed scaled pattern loss, and a dynamic masked cycle-consistent auxiliary network.
Explicit information that could indicate the position of defects is intentionally excluded to avoid learning any direct mapping.
Experimental results on the texture class of the challenging MVTec AD dataset show that the proposed method is 2.9\% higher than the SOTA methods in F1-Score, while substantially outperforming SOTA methods in generalizability.
\end{abstract}


\section{Introduction}
\label{sec:intro}
Defect detection is a crucial and formidable task in industrial production and automation.
Defects exhibit uncertain appearances, often appearing as small anomalies that are partially intermixed with normal regions.
The cost of manual annotation and the difficulty of collecting representative samples limit the effectiveness of conventional visual inspection methods, leading recent approaches to adopting unsupervised techniques for addressing this task.

In recent years, the rapid development of generative methods has led to the proposal of several reconstruction-based methods.
Many methods employ Generative Adversarial Network (GAN) \cite{goodfellow2020generative} or Autoencoder \cite{bergmann2018improving} for image reconstruction, followed by anomaly level estimation through comparison of the input and the reconstructed images.
The aforementioned methods, however, suffer from imperfect reconstructions and noisy outputs resulting from suboptimal distance metrics.
Despite attempts \cite{akcay2018ganomaly,akccay2019skip,deng2022anomaly,zavrtanik2021draem}  to address these issues, they share a similar paradigm: identifying defects by modeling the normal distribution.


However, the training source may not provide sufficient data to adequately model normality, leading to imperfect reconstructions. 
Variations in environmental factors could also disrupt the distribution of normality, constituting difficulty for generalizing.
The purpose of using distance metrics in previous methods is to identify the regions that have undergone certain modifications while being restored to the modeled normality, which also no longer applies if normality changes.
Despite these limitations, it appears theoretically infeasible to detect anomalies in object classes without referencing the normal distribution.
The main reason is that their standards of normality heavily rely on human definition, which varies among different products.
The modeling process upon the given normal samples can be seen as learning the anomaly standard of that particular class.
This standard is not transferable across object classes since the same type of anomaly (\eg rotation) may not be considered anomalous in another class.
Texture classes, however, share the same anomaly standard, making the modeling process redundant as defects can be discerned by examining the samples alone.
We aim to directly localize the defective areas without any intermediate process that models the distribution of the normal samples, thereby avoiding the side effects caused by normality modeling.
We commence by examining simple self-supervised methods and promptly discern that they may adopt shortcuts due to the limited number of training samples.
This has prompted us to adopt an innovative approach that eliminates all ground truth information in the learning process, thereby precluding any shortcuts.

Specifically, we employ a Generative Adversarial Network (GAN) to find the internal relationship between synthetic anomalies and arbitrary masks, while emphasizing that these two distributions are entirely independent.
The generator is expected to find correlations between them and produce annotations relevant to the input.
The correlation can be simplified as follows: an annotation mask is characterized by distinct black and white markings, representing its binary nature. 
Similarly, the generator must recognize that the inputs also possess a binary nature, consisting of both defective and normal regions.
Subsequently, it can acquire the ability to associate negative areas with black markings and vice versa.
The outputs initially conformed to our expectations, however, the network subsequently started generating arbitrary masks since the training objective did not specifically require relevant outputs.

To facilitate the discovery of the correlation between the two distributions, we propose a technique termed ``Invisible Pattern'' that conceals pattern information exclusively within the semantic negative area, accompanied by an innovative loss function called the scaled pattern loss.
It establishes a bidirectional pixel-wise mapping, thereby preventing the generator from producing arbitrary outputs.
Inspired by previous methods \cite{zhu2017unpaired}, we apply a cycle-consistent regulation to ensure that the prediction encodes the essential information, and successfully stabilizes the generation of the invisible pattern.
Additionally, a dynamic correction mechanism is also designed to incentivize the network to localize defects more accurately in the absence of ground truth.
This innovative approach allows the network to refine its predictions based on its own predictions instead of depending on corrections from external sources.
Through the collaborative efforts of these components, the network is capable of accurately identifying and assigning appropriate annotations to their corresponding areas in the input.

In contrast to previous approaches specifically designed for modeling normality distribution or discriminative learning mappings based on ground truth masks, our method is trained without explicit guidance to avoid leaking information that could indicate the location of defects and prevent overfitting.
We provide unrelated masks solely just to ensure that the outputs resemble a mask while allowing the network to autonomously learn how to accurately localize defects.
Our approach has successfully trained a model that operates seamlessly without the need for intermediate processes, showcasing robust generalization capabilities not only on test sets but also on additional test samples with diverse environmental variations.
Experiments also suggest that the performance of our method is superior to the previous state-of-the-art methods while operating with no requirement for a threshold process before actual deployment.
During the inference stage, our framework requires only one forward pass in the main generator, resulting in a twofold increase in operational speed compared to state-of-the-art methods.

Overall, our main contributions are listed as follows:
\begin{itemize} 
\item We propose a novel approach that leverages correlations between unrelated samples and masks to acquire the capability of defect localization.
\item We propose a scaled pattern loss that enables a bidirectional pixel-wise mapping between the input and the output.
\item We propose a dynamic masked cycle-consistent structure to implicitly optimize the localization results.
\end{itemize}

\begin{figure*}[htbp!]
    \centering
	\includegraphics[width=\linewidth]{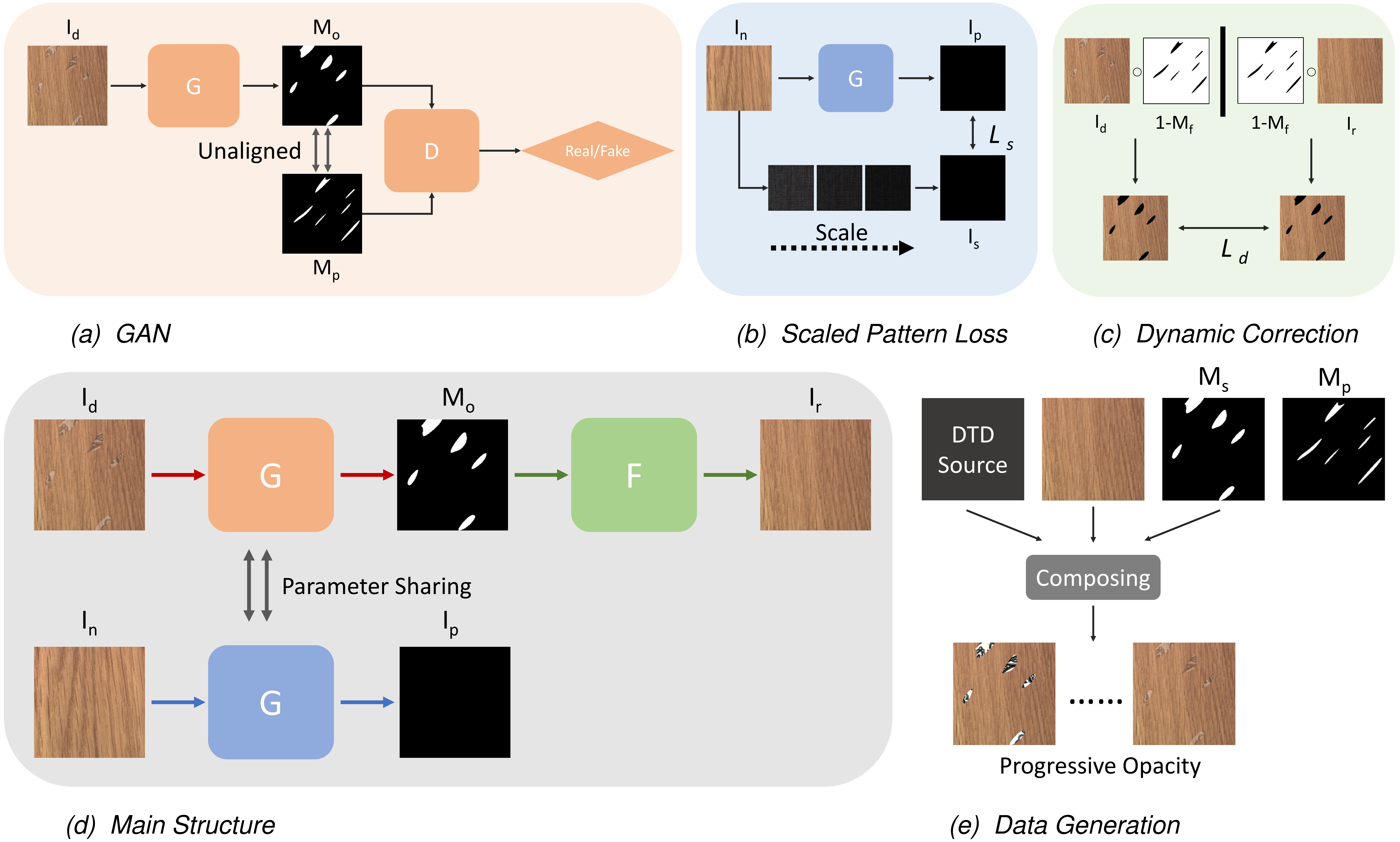}
      \begin{subfigure}{0.41\textwidth}
          \phantomcaption
          \label{fig: framework a}
      \end{subfigure}
      \begin{subfigure}{0.21\textwidth}
          \phantomcaption
          \label{fig: framework b}
      \end{subfigure}
	\begin{subfigure}{0.21\textwidth}
          \phantomcaption
          \label{fig: framework c}
      \end{subfigure}
	\begin{subfigure}{0.5\textwidth}
          \phantomcaption
          \label{fig: framework d}
      \end{subfigure}
	\begin{subfigure}{0.21\textwidth}
          \phantomcaption
          \label{fig: framework e}
      \end{subfigure}
\caption{
\label{fig: framework}%
An overview of the proposed framework.
In the training stage, the above components jointly train \(G\) to adopt the ability of defect detection.
In the inference stage, we forward the inputs through \(G\) to acquire the localization results.}
\end{figure*}


\section{Related work}
\label{sec: related work}
We briefly review previous literature on surface defect detection in this section.
A primary difficulty in tackling this problem is the acquisition of defective samples.
Consequently, recent literature predominantly emphasizes unsupervised or self-supervised methodologies.

A great number of anomaly detection models adopted a reconstructive approach.
Generative methods such as GAN \cite{schlegl2017unsupervised,schlegl2019f} or autoencoders \cite{bergmann2018improving,gong2019memorizing} enable powerful reconstruction ability using normal data only.
Under the assumption that these models would not successfully recover the defects, the localization map is produced by the reconstruction error between the image and its reconstruction \cite{sakurada2014anomaly,zavrtanik2021reconstruction}.
Distance metrics such as \(l_2\) distance \cite{hadsell2006dimensionality} or SSIM \cite{bergmann2018improving} are used for better measurement.
However, the models might be able to reconstruct defects as well though they were not present in training, leading to low error in the comparison and causing inaccurate localization results.
In \cite{zavrtanik2021draem}, a discriminative network is introduced to capture subtle differences between the inputs and the reconstructions.
Some other works have tried to use memory-based autoencoders to further avoid defective reconstructions \cite{park2020learning}.

Besides the over-generalization of defects, another issue is the unsatisfied reconstructions regarding normal regions, leading to inaccurate localization.
Some methods seek ways to generate images of higher quality. 
\citeauthor{bergmann2018improving} apply structural similarity to autoencoders and improve the reconstruction quality.
In \cite{deng2022anomaly}, a trainable one-class bottleneck embedding module is designed to acquire more accurate reconstructions while excluding noises.


Due to the lack of defect samples, which is essential for the training process of a lot of frameworks, self-supervised methods are introduced to serve as a solution to the dire need for defect samples.
Many methods \cite{devries2017improved,yun2019cutmix,zhong2020random} fabricate anomalies by replacing small rectangular regions of the original image with other values.
\cite{li2021cutpaste} propose the CutPaste augmentation to generate defect images by cutting a structural image patch from itself and randomly pasting them to other places.
It highlights the need to synthesize samples close to the normal image distribution.
\citeauthor{zavrtanik2021draem} proposed an augmentation method by utilizing a Perlin noise mask to combine normal images with random image sources from \cite{cimpoi2014describing}.
The generated anomalies take various forms and contain both subtle and obvious defects. 

These methods provide solid solutions for data augmentation.
However, a gap exists between the distributions of the fabricated data and real-world defects.
This results in generalization issues and severely impacts the performance of self-supervised methods.
While the corrections provided by labels are proved to be useful in training, overfitting issues could occur at the same time.

\section{Method}
\label{sec: method}
In this paper, we propose a novel approach for defect localization that significantly diverges from previous methods.
The presented method, as shown in \cref{fig: framework d}, is a dual-path framework with two distinct inputs (normal and defective). 
Both paths share a common generator \(G\) with identical parameters, updated concurrently through feedback from both routes.
The fundamental concept is to deliberately exclude explicit information, enabling the network to learn to localize anomalies in textures rather than merely memorizing local appearances without considering the entire image.
To train in the absence of real-world defective samples, we gain inspiration from the augmentation method proposed in \cite{zavrtanik2021draem} along with a newly proposed progressive opacity strategy to create the training samples.

\subsection{Unpaired Transition}
\label{sec: Contrastive Localization}
As shown in \cref{fig: framework a}, the first step is to establish a GAN structure, consisting of a generator \(G\) and a discriminator  \(D_Y\).
The objective is formulated as:
\begin{align} \label{eq:1}
\mathcal{L}_{GAN}(G,D_Y,X,Y)	= \; &\mathbb{E}[log(1-D_Y(G(x))]
								\nonumber \\
						+ \; &\mathbb{E}[logD_Y(y)],
\end{align}
where \(X\) and \(Y\) represent the domain of defective samples and random masks respectively, and \(x$ and $y\) are the samples stochastically drawn from each domain. 
Both \(X\) and \(Y\) are generated in \cref{sec: data}, but they are not related which means \(y\) does not possess the correct information that could localize defects in \(x\).

The generator is fed with defective samples \(I_d\) to generate an output \(M_o\) that simulates a randomly generated mask \(M_p\).
Typically, the output \(M_o\) in this setup should be a random mask.
However, even without ground truth information, GAN can establish weak associations between defective inputs and random annotations, although the outcome is unstable and quickly deteriorates into random masks.
The generator has been conditioned with the defective input, and we expect it to generate a mask based on its global observation of the condition.
Even without feedback from the correct mask, the generator can still retain certain information from the data characteristics of the defective inputs.
We hope to exploit this trait to facilitate the generator in acquiring the capability of transforming the defective segment within an image into corresponding localization masks, without any explicit guidance.
Since this connection is weak, unstable, and lacks precision, we propose various constraints and catalysts to stabilize and enforce it to achieve precise localization of defects. 
Even though we already possess the corresponding self-supervised ground truth \(M_s\), we exclude the possibility of using self-supervised approaches to achieve a possibly more stable and precise transition in training.
The reliance it brings to the generator on the corresponding ground truth causes information leaks and overfitting, which leads to shallow mapping and undesirable performance in testing.

\subsection{Scaled Pattern Loss}
In this section, we aim to stabilize the GAN transition process, enabling the generator to effectively navigate between two distributions instead of generating random masks that merely satisfy the training objective.
Therefore, to establish a relatively stable connection between the inputs and outputs, we introduce a concept called the invisible pattern and propose a new loss function termed the scaled pattern loss (\(SP\) loss).
As shown in \cref{fig: framework b}, we create a new input path of normal images \(I_n\) and urge the network to convert it into the corresponding invisible pattern.
Note that \(I_p\) is expected to be a linearly scaled version of the input, not an empty mask.
The goal of the \(SP\) loss is to create a pixel-wise bidirectional mapping between the input and the output of the generator, preventing the generator from producing arbitrary localizations.
Since the invisible pattern is imperceptible to the human eye, its corresponding regions can be directly interpreted as negative predictions during inference.
The formula is shown below:
\begin{align}
\label{eqn: scaled}
I_{s}=I_{n}\times\alpha-\beta\\
\mathcal{L}_{S}(I_{p},I_{s})= \;||I_{p}-I_{s}||_1,
\end{align}
where \(I_{s}\) is the scaled version of inputs and  \(I_{p}\) is the output for normal samples.
The symbols \(\alpha\) and \(\beta\) control the scale level of the image.
Both of them are adjustable hyperparameters, and we recommend setting \(\alpha=0.005\) and \(\beta=0.995\).

Despite being trained solely on normal images, the second path exerts a simultaneous influence on the outputs for anomalous images, thereby effectively preserving texture information within its region of negative predictions.
The invisible pattern, distributed in a data range close to zero, seamlessly integrates with the original zero values in the generator output, enforcing it to generate outputs pixel-wise correlated with the input.
However, this proximity also means that if the network produces zero values instead of the invisible pattern in these regions, the loss will still be low.

\subsection{Dynamic Correction}
\label{auxiliary}
Despite the theoretical feasibility of the above structure, the training process is found to be volatile in practice. 
As training progresses, the adversarial loss and the scaled pattern loss often collide and fail to coexist, with one potentially dominating and excluding the effect of the other.
Since the target \(M_p\) is a mask consisting only of values zero and one, the dominance of the discriminator will erase the invisible pattern, replacing them solely with zero values. 
Conversely, if the \(SP\) loss becomes predominant, \(G\) will convert all inputs into the invisible pattern, thereby impeding the generation of the positive mask.

Inspired by previous works\cite{zhu2017unpaired}, we adopt the encoder-decoder paradigm behind it and perceive the prediction of the generator as an encoded version of the input.
Specifically, we apply a cycle-consistent structure that consists of a second generator \(F\) and a corresponding discriminator \(D_X\).
Prediction \(M_o\) is processed by \(F\) to retrieve the original input from the invisible pattern, and \(D_X\) discriminates the recovered image \(I_r\) with \(I_d\).
The demand for pattern information drives the generation of the invisible pattern, improving overall network stability.
However, we observed a slight decline in performance and promptly recognized that this is attributed to the unsuccessful recovery of defective areas.
The presence of these regions hampers further reduction in loss and induces network turbulence, thereby impeding network convergence.

A direct solution is to utilize the mask \(M_s\) to exclude the defective region during the loss calculation process, but it leaks the defect location to the network.
To avoid reliance on ground truth information, we propose a dynamic correction mechanism that stabilizes the loss and improves localization precision without requiring knowledge of defect locations.
Precisely, we utilize the prediction from generator \(G\) to mask the defective regions in the inputs and the corresponding area in the recovered image \(I_r\) as in \cref{fig: framework c}.
As localization accuracy improves, the turbulence from the defective region will gradually diminish, enhancing prediction quality. 
This establishes a positive feedback loop.
Conversely, omissions of defects will lead to comparisons of unsuccessfully recovered areas, increasing the \(DMCC\) loss.
False positives in predictions result in the occurrence of positive masks in the normal path, thereby increasing the \(SP\) loss.
Together, these modules impose various constraints, allowing the generative adversarial network to leverage itself to identify relationships between uncorrelated distributions for precise defect localization.
This process is denoted as \(M\) and the objective can be formulated as:
\begin{align}
\mathcal{L}_{cyc}(G,F)	= \; &\mathbb{E}[||M(F(G(x))-M(x)||_1]
								\nonumber \\
						+ \; &\mathbb{E}[||G(F(y)-y||_1].
\end{align}

To maintain the balance between the losses, we assign proper weight to each loss and the full objective is:
\begin{align}
\label{eqn: all}
\mathcal{L}(G,F,D_X,D_Y)	&= \mathcal{L}_{GAN}(G,D_Y,X,Y)\nonumber\\
					&+ \mathcal{L}_{GAN}(F,D_X,Y,X)\nonumber\\
					&+ \lambda_1\mathcal{L}_{cyc}(G,F)\nonumber\\
					&+ \lambda_2\mathcal{L}_{S}(I_{p},I_{s}).
\end{align}
where \(\lambda_1\) and \(\lambda_2\) are set to 10 and 0.4 respectively, while considering adjustments based on the overall balance of the network.

\subsection{Data Augmentation}
\label{sec: data}
In training, our method requires the presence of both defective and normal images.
With only normal samples being accessible, augmentation methods could serve as an alternative to provide the defect samples needed. 
We adopt the synthesis technique proposed by \cite{zavrtanik2021draem} to generate defective samples while utilizing another Perlin noise mask \(M_p\) generated during this process as the example for the discriminator.
Note that, \(M_p\) are unaligned with \(M_s\) which is the mask that synthesis defects, which means no paired samples are utilized during training.
Given the unique training method, we conjecture that the final performance of the model will be heavily influenced by the quality of the simulated samples.
Transparency, in particular, plays a key role, and a model trained with high-transparency samples is more sensitive to subtle defects. 
However, introducing more transparent defects results in unstable training due to potential failure in perceiving defective areas in the early training stage.

To address this, we propose a progressive opacity strategy (\(Pos\)) that gradually increases transparency throughout the training process, ultimately yielding nearly imperceptible defects.
The training data is regenerated every 50 epochs based on a progressive lower opacity value provided.
With \(Pos\) instead of random opacity, the model could reach a stable state faster by initially using apparent defects, and it is optimized to detect more subtle defects later.

\section{Experiments}
\label{sec:experiments}
In this section, we benchmark the proposed method on a diverse set of texture defect detection datasets.
The generalizability of the proposed method is also evaluated under simulated environment conditions.


\subsection{Benchmarks}
We compare our method qualitatively with recent works CFlow \cite{gudovskiy2022cflow}, DRAEM  \cite{zavrtanik2021draem}, PatchCore  \cite{roth2022towards}, and PyramidFlow \cite{lei2023pyramidflow}.

{\bf Datasets.} The MVTec anomaly detection dataset \cite{bergmann2019mvtec} consists of 5 texture classes and 10 object classes. 
Our proposed method is specifically designed for detecting texture defects, thus the 5 texture classes are utilized in the evaluation.
To further enrich the test source, we also benchmark on the DTD-Synthetic constructed by \cite{aota2023zero} and the Woven Fabric Textures proposed in \cite{bergmann2018improving}.

\begin{figure}[htb]
	\centering{
		\includegraphics[width=\linewidth]{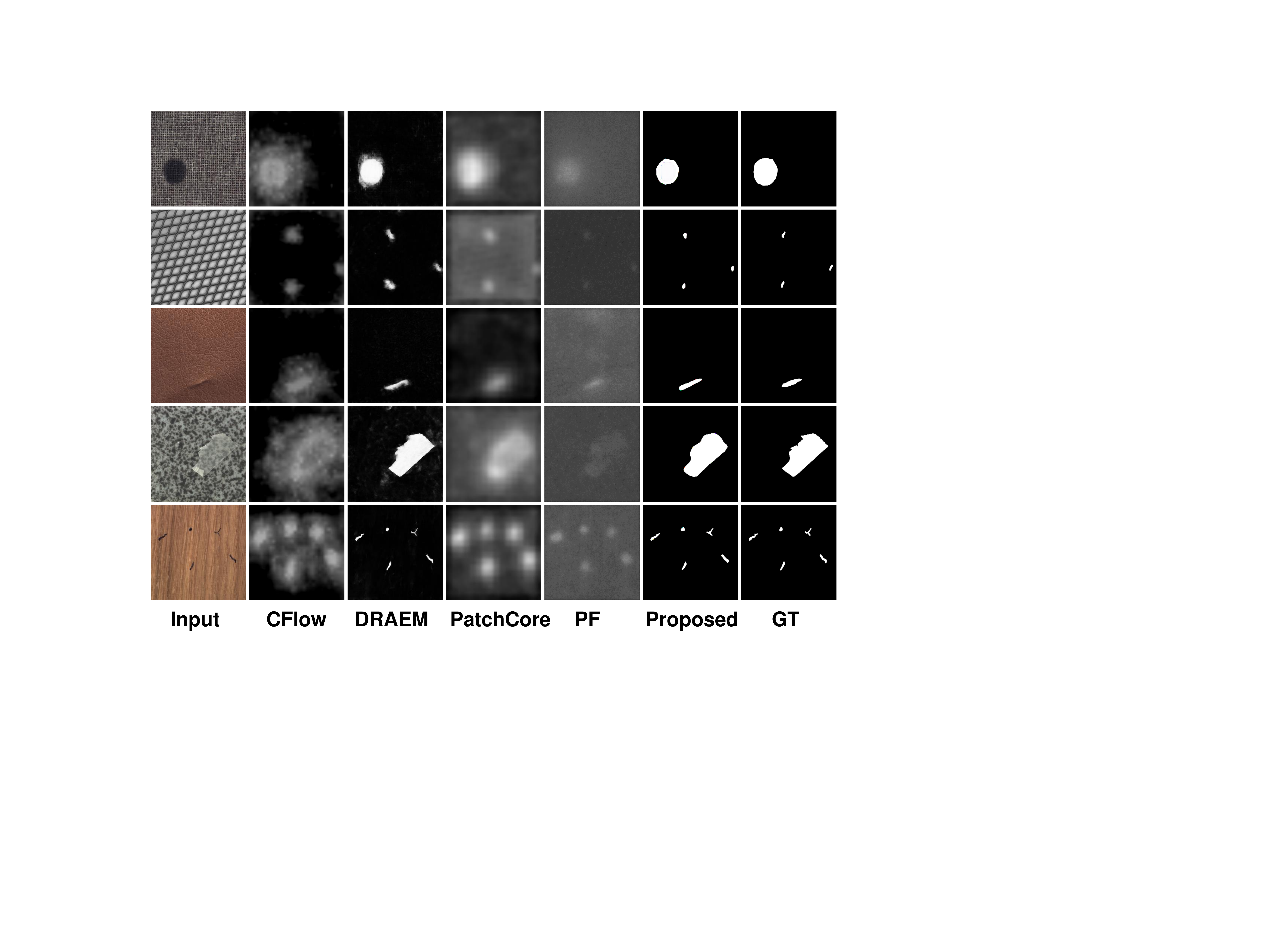}
		}
	  \caption{\label{fig: comparison}Comparison of anomaly maps (No threshold). The anomalous images and the ground truth are shown in the first and the last column. The middle five columns, from left to right, are results from CFlow \cite{gudovskiy2022cflow}, DRAEM \cite{zavrtanik2021draem}, PatchCore   \cite{roth2022towards}, PyramidFlow \cite{lei2023pyramidflow} and our proposed method.}
\end{figure}

{\bf Metrics.} AUROC is a commonly used metric that measures performance across all thresholds.
However, in imbalanced problem settings, it tends to disregard false positives and excessively penalize false negatives.
Localizations in defect detection should not be excessively ambiguous, as it is crucial to acknowledge that classification tasks already the presence of defects in specific regions.
Methods with vague localization results often achieve an AUROC near 1 (the maximum value), while offering obvious deficiencies in their results as shown in \cref{fig: comparison}.
F1-Score is a metric that is specifically designed for imbalanced problem settings, making it a more suitable evaluation metric in the experiments below.

 
{\bf Experimental settings.} The images are resized to a predetermined resolution ( \(256 \times 256\) ) and then normalized being fed into the network.
The generator \cite{he2016deep} and discriminator \cite{isola2017image} adopt structures that are similar to previous methods \cite{zhu2017unpaired, kim2019u}. 
The model is trained for 1000 epochs using a batch size of 1 on an RTX 3080 GPU.
An Adam optimizer \cite{kingma2014adam} with \(\beta = (0.5, 0.999)\) is applied with a 0.0002 learning rate.
The data augmentation method mentioned in \cref{sec: data} is utilized to generate a total of 300 simulated defective samples and masks for training.

\begin{table}[htb]
\centering
\resizebox{\linewidth}{!}{
\begin{tabular}{l *{5}{c}}
\toprule
Class&CFlow \cite{gudovskiy2022cflow}&DRAEM \cite{zavrtanik2021draem}&PatchCore  \cite{roth2022towards}&PyramidFlow\cite{lei2023pyramidflow}&Ours\\
\midrule
\midrule
	   Wood & 53.5  & 68.3  & 47.2  & 48.6  & \textbf{70.7}   \\ 
        Carpet & 63.6  & 50.5  & 59.8  & 54.1  & \textbf{66.1}   \\ 
        Tile & 66.8  & \textbf{91.3}  & 63.6  & 70.6  & 86.2   \\ 
        Leather & 55.4  & 63.5  & 45.0  & 46.4  & \textbf{66.4}   \\ 
        Grid & 32.4  & \textbf{56.8}  & 18.5  & 32.9  & 55.8   \\ 
        AVG & 54.3  & 66.0  & 46.8  & 50.5  & \textbf{68.9}   \\ 
\bottomrule
\end{tabular}
}
\caption{
The F1-Score of the anomaly localization results on the MVTec \cite{bergmann2019mvtec} dataset
}
\label{tab: performance}
\end{table}%

\begin{table}[htb]
\centering
\resizebox{\linewidth}{!}{
\begin{tabular}{l *{5}{c}}
\toprule
Class&CFlow \cite{gudovskiy2022cflow}&DRAEM \cite{zavrtanik2021draem}&PatchCore  \cite{roth2022towards}&PyramidFlow\cite{lei2023pyramidflow}&Ours\\
\midrule
\midrule
       Texture\_1 & 65.5  & 56.0  & \textbf{71.7}  & 37.3  & 62.6   \\ 
        Texture\_2 & 66.9  & 66.4  &\textbf{ 67.8 } & 39.2  & 52.1   \\ 
        Blotchy\_099 & 71.1  & 78.0  & 62.3  & 46.2  &\textbf{ 85.9 }  \\ 
        Fibrous\_183 & 61.1  & 76.2  & 52.1  & 43.9  & \textbf{84.3 }  \\ 
        Marbled\_078 & 65.1  & 72.1  & 59.3  & 46.1  & \textbf{78.4 }  \\ 
        Matted\_069 & 66.5  & 57.4  & 55.2  & 52.8  & \textbf{78.8 }  \\ 
        Mesh\_114 & 43.4  &\textbf{ 64.4}  & 41.7  & 23.1  & 62.1   \\ 
        Perforated\_037 & 30.5  &\textbf{ 69.4 } & 45.5  & 22.4  & 64.3   \\ 
        Stratified\_154 & 61.4  & \textbf{75.4 } & 52.1  & 54.3  & 73.4   \\ 
        Woven\_001 & 53.6  & 76.7  & 42.2  & 40.6  & \textbf{78.5}  \\ 
        Woven\_068 & 55.0  & 52.4  & 45.1  &\textbf{ 66.9 } & 61.2   \\ 
        Woven\_104 & 49.7  &\textbf{ 53.7 } & 52.4  & 15.6  & 47.1   \\ 
        Woven\_125 & 68.3  & 68.1  & 59.7  & 51.3  &\textbf{ 76.5 }  \\ 
        Woven\_127 & 28.3  & 29.1  &\textbf{ 58.3}  & 43.4  & 28.6   \\ 
        AVG & 55.8  & 64.0  & 56.7  & 41.7  & \textbf{66.7 }  \\ 
\bottomrule
\end{tabular}
}
\caption{
The F1-Score of the anomaly localization results on the DTD-Synthetic \cite{aota2023zero} and the Woven Fabric Textures \cite{bergmann2018improving}.
}
\label{tab: performance2}
\end{table}%

{\bf Performance.} The performance of our proposed method is compared with state-of-the-art approaches in \cref{tab: performance}. 
The best results are highlighted in boldface.
Our method outperforms the previous SOTA by 2.9\% on the MVTec dataset and by 2.7\% on the combination of the DTD-Synthetic and Woven Fabric Textures.
A visual comparison with previous methods is presented in \cref{fig: comparison}, and our method provides precise localizations with less noise.
The average prediction time of DRAEM \cite{zavrtanik2021draem} is 0.013 seconds, whereas our method achieves a prediction time of approximately 0.006 seconds under the same test environment, doubling its speed.
Further details on the evaluation of inference speed and additional experiments conducted with additional datasets are provided in the supplementary material.


\begin{table*}[htb]
\centering
\resizebox{\linewidth}{!}{
\begin{tabular}{l*{12}{c}}
\toprule
      &\multicolumn{5}{c}{General Setting}&\multicolumn{6}{c}{Hard Setting}\\
\cmidrule(lr){2-6}\cmidrule(lr){8-12}
Class&CFlow \cite{gudovskiy2022cflow}&DRAEM \cite{zavrtanik2021draem}&PatchCore  \cite{roth2022towards}&PyramidFlow\cite{lei2023pyramidflow}&Ours& &CFlow \cite{gudovskiy2022cflow}&DRAEM \cite{zavrtanik2021draem}&PatchCore  \cite{roth2022towards}&PyramidFlow\cite{lei2023pyramidflow}&Ours\\
\midrule
        Wood & 46.7  & 11.9  & 44.1  & 11.6  & \textbf{62.7} &  & 3.8  & 7.8  & 14.2  & 9.4  & \textbf{50.0}  \\ 
        Carpet & 56.0  & 5.3  & 51.2  & 5.2  & \textbf{63.2} &  & 51.6  & 3.5  & 52.3  & 11.0  & \textbf{57.2}  \\ 
        Tile & 54.5  & 15.7  & 61.3  & 21.1  & \textbf{75.4} &  & 54.6  & 14.3  & 55.6  & 36.2  & \textbf{63.5}  \\ 
        Leather & 47.8  & 1.6  & 44.9  & 12.4  & \textbf{49.8} &  & 9.3  & 1.3  & 19.3  & 1.9  & \textbf{29.9}  \\ 
        Grid & 31.5  & 2.4  & 7.3  & 2.6  & \textbf{47.7} &  & 31.2  & 2.0  & 8.9  & 3.3  & \textbf{48.2}  \\ 
\bottomrule
\end{tabular}
}
\caption{Results for generalization experiments under general setting (left) and hard setting (right) on  CFlow \cite{gudovskiy2022cflow}, DRAEM \cite{zavrtanik2021draem}, PatchCore  \cite{roth2022towards}, PyramidFlow\cite{lei2023pyramidflow} and our proposed method.}
\label{tab: generalization}
\end{table*}%

\subsection{Generalizability}
\label{sec: gen}
Previous approaches have demonstrated consistent advancements in achieving optimal performance.
The generalizability of the trained model, regarding its accuracy on test samples with different environmental conditions, has not been given enough attention.
In the real world, texture appearance can vary significantly due to factors such as lighting and chromatism, making it crucial to assess generalizability in unseen scenarios for determining model applicability.

\begin{figure}[htb]
	\centering{
		\includegraphics[width=\linewidth]{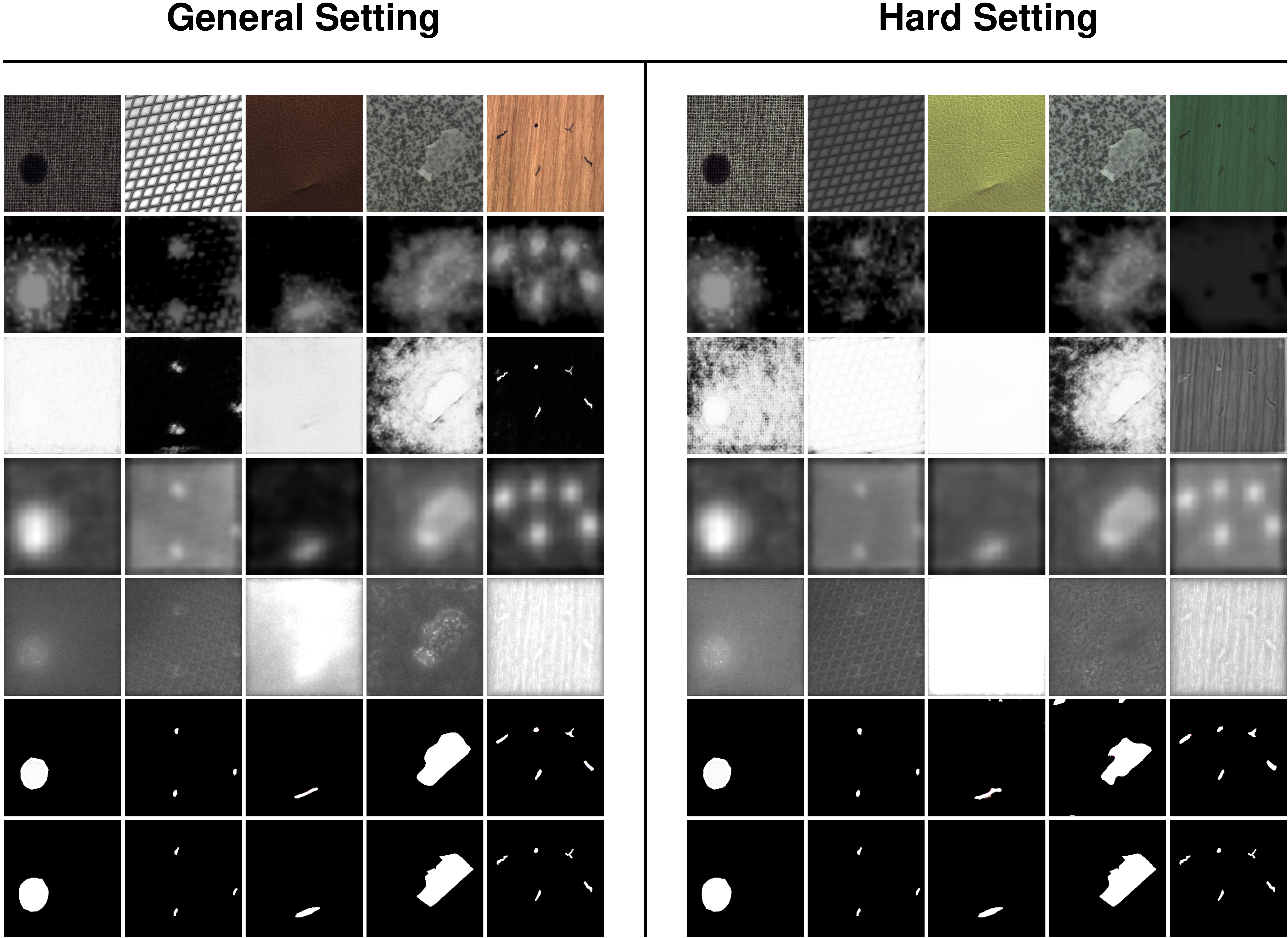}
		}
	  \caption{\label{fig: generalization}Localization results on generalization ability. The test source consisting of the general setting (left) and the hard setting (right) is shown in the first row. The ground truth is at the bottom of the figure. The middle five rows show the localization results from CFlow \cite{gudovskiy2022cflow}, DRAEM \cite{zavrtanik2021draem}, PatchCore   \cite{roth2022towards}, PyramidFlow \cite{lei2023pyramidflow} and our proposed method respectively. }
\end{figure}

In this subsection, we conduct additional experiments in simulated environments to evaluate the generalizability of our approach.
Image attributes, such as Contrast, Brightness, and Hue are utilized to create a simulated illumination environment of two levels of settings.
The general setting adjusts both brightness and contrast, whereas the hard setting additionally modifies hue.
As shown in \cref{fig: generalization} and \cref{fig: comparison}, most methods are affected to some extent in the general setting, and the hue changes further exacerbate performance degradation. DRAEM\cite{zavrtanik2021draem} appears to be significantly impacted by the simulated environments, as it experiences significant drops in performance in both settings.
The statistics presented in \cref{tab: generalization} indicate that previous methods all experience significant performance drops in a few classes under unseen test scenarios, whereas our method is capable of maintaining certain performance and outperforming others by \textbf{12.4\%} and \textbf{18.3\%} in the general and hard settings, respectively.


DRAEM \cite{zavrtanik2021draem} exhibited the most significant decrease in the experiments. 
After further inspections, we believe it is a shared issue of reconstruction-based methods.
In their reconstruction process, they transform the data from the anomaly distribution to a normal distribution that is modeled in the training process.
However, this normal distribution is quite shallow and only varies within the approximate data range that has been given in training.
In this case, a simple change in illumination intensity would shift the correct reconstructions of the test samples out of the modeled distribution, causing it to misfunction.
For example, a darker anomaly sample will still be reconstructed into a normal sample with the original brightness, and it will cause a lot of differences between them and result in many false positives. 
A More detailed analysis is provided in the supplementary material.
As for feature-based methods like PaDim \cite{defard2021padim} and PatchCore \cite{roth2022towards}, they are less affected by this issue as high-level features contain less information about the illumination and color changes on the inputs. 
However, although they are not as severely affected as in reconstruction-based methods, we can observe from \cref{tab: generalization} that the more changes are introduced into the test scenario, the more apparent the effect becomes and the performance for a few classes drops to a point where their predictions become meaningless.
The two approaches both rely on modeling the normal distribution and referencing them to make predictions. 
The first one learns to transform defective inputs into normal samples, while the second one learns a distribution of normality in the feature space.
In unseen test scenarios, changes in normality occur, making the normal distribution learned during training invalid.

In contrast to existing methods that reference normality, our approach localizes defective regions on the given samples directly.
In training, there is no explicit information to inform the network where is the defect and how to localize it.
Our framework is trained without the presence of ground truth, and there are no manually designed procedures that are used for determining the existence of a defect, such as comparing feature similarity with normal samples or measuring reconstruction differences.
Therefore, the network will not associate the prediction with certain appearances as specifically shown in \cref{fig: generalization}.
There is also no designed procedure to enforce it to reference a modeled distribution to make predictions.
Our network learns to localize defects by continuously attempting to make predictions and being corrected by the dynamic correction mechanism.
In this process, we believe the network learns to identify the semantics of a defect based on a global observation of the inputs.
Since the inference does not rely on a fixed normality, the network avoids assumptions about the input distribution, thereby demonstrating improved robustness in adapting to environmental variations.
Although our method exhibits a certain decline in unseen test scenarios, its performance maintains an acceptable level with an F1-Score of 59.7\% in the general setting and 49.7\% in the hard setting.

\section{Ablation study}
In this section, we perform an ablation analysis to investigate the individual contributions of each module and the impacts of data choice, which is essential for comprehending our framework.




\begin{table*}[htb]
\centering
\begin{tabular}{l*{10}{c}}
\toprule
    &\multicolumn{4}{c}{Architecture}&\multicolumn{5}{c}{Opacity}\\
\cmidrule(lr){2-5}\cmidrule(lr){6-10}
Method&GAN&Scaled&Cyc&Mask&0(0.1)&0(0.5)&0(0.9)&O(0.1-0.9)&Pos&AVG\\
\midrule
G&\checkmark&&&&&&&\checkmark&&2.8\\
G.S&\checkmark&\checkmark&&&&&&\checkmark&&37.2\\
G.C.M&\checkmark&&\checkmark&&&&&\checkmark&&27.4\\
G.S.C&\checkmark&\checkmark&\checkmark&&&&&\checkmark&&35.4\\
G.S.C.M&\checkmark&\checkmark&\checkmark&\checkmark&&&&\checkmark&&51.1\\
G.S.C.M(0.1)&\checkmark&\checkmark&\checkmark&\checkmark&\checkmark&&&&&15.9\\
G.S.C.M(0.5)&\checkmark&\checkmark&\checkmark&\checkmark&&\checkmark&&&&40.4\\
G.S.C.M(0.9)&\checkmark&\checkmark&\checkmark&\checkmark&&&\checkmark&&&30.9\\
G.S.C.M(Pos)&\checkmark&\checkmark&\checkmark&\checkmark&&&&&\checkmark&\textbf{68.9}\\
\bottomrule
\end{tabular}
\caption{Anomaly localization results of ablation study on architecture (left) and opacity (right).}
\label{tab: ablation}
\end{table*}%

\subsection{Architecture}
The components are reassembled to form different structures.
These components consist: 
(\textit{\RNum{1}}) GAN represents only the basic GAN structure
(\textit{\RNum{2}}) Scaled denotes the scaled pattern loss with another path of normal inputs.
(\textit{\RNum{3}}) Cyc is the auxiliary network with an additional generator and discriminator, enabling cycle-consistent loss.
(\textit{\RNum{4}}) Mask refers to the dynamic correction mechanism.
(\textit{\RNum{5}}) Pos is the progressive opacity strategy that we applied in data generation.
A combination of initials, such as \(G.S.C\), represents the enabled components and the content within the parentheses indicates the opacity setting.

\cref{tab: ablation} shows that \(G.S.C.M(Pos)\) demonstrates optimal performance compared to others, while the simplest form \(G\) only produces a result of 2.8\%.
Upon closer inspection, the table reveals a general increasing trend as each component is added.
Contrary to the overall trend, the performance of \(G.S.C\) has deteriorated with the addition of the Cyc component.
In \cref{auxiliary}, we briefly discussed the potential reasons for this behavior, and one of the aims of the dynamic correction mechanism is to address this issue as well.
Obviously, \(G\) alone is not capable of learning to make valid predictions.
With additional components, whether in \(G.S\) or \(G.C.M\), the prediction reaches a functional point.
In \(G.S.C.M\) and \(G.S.C.M(Pos)\), there are two more major rises in the performance.
However, unlike other methods, these components are not part of a linear process, and their joint effort falls onto the generator alone.
This signifies that these components serve different purposes, as simply stacking components with similar purposes will not result in a significant performance increase each time.
\label{ablation: aug}
\subsection{Data Source}
In this framework, the model is not guided by any explicit information and needs to navigate the correct way of localizing defects.
The opacity value significantly affects defect visibility and plays a crucial role in determining recognition difficulty, making it a pivotal training factor.

\begin{figure}[htb]
	\centering{
		\includegraphics[width=\linewidth]{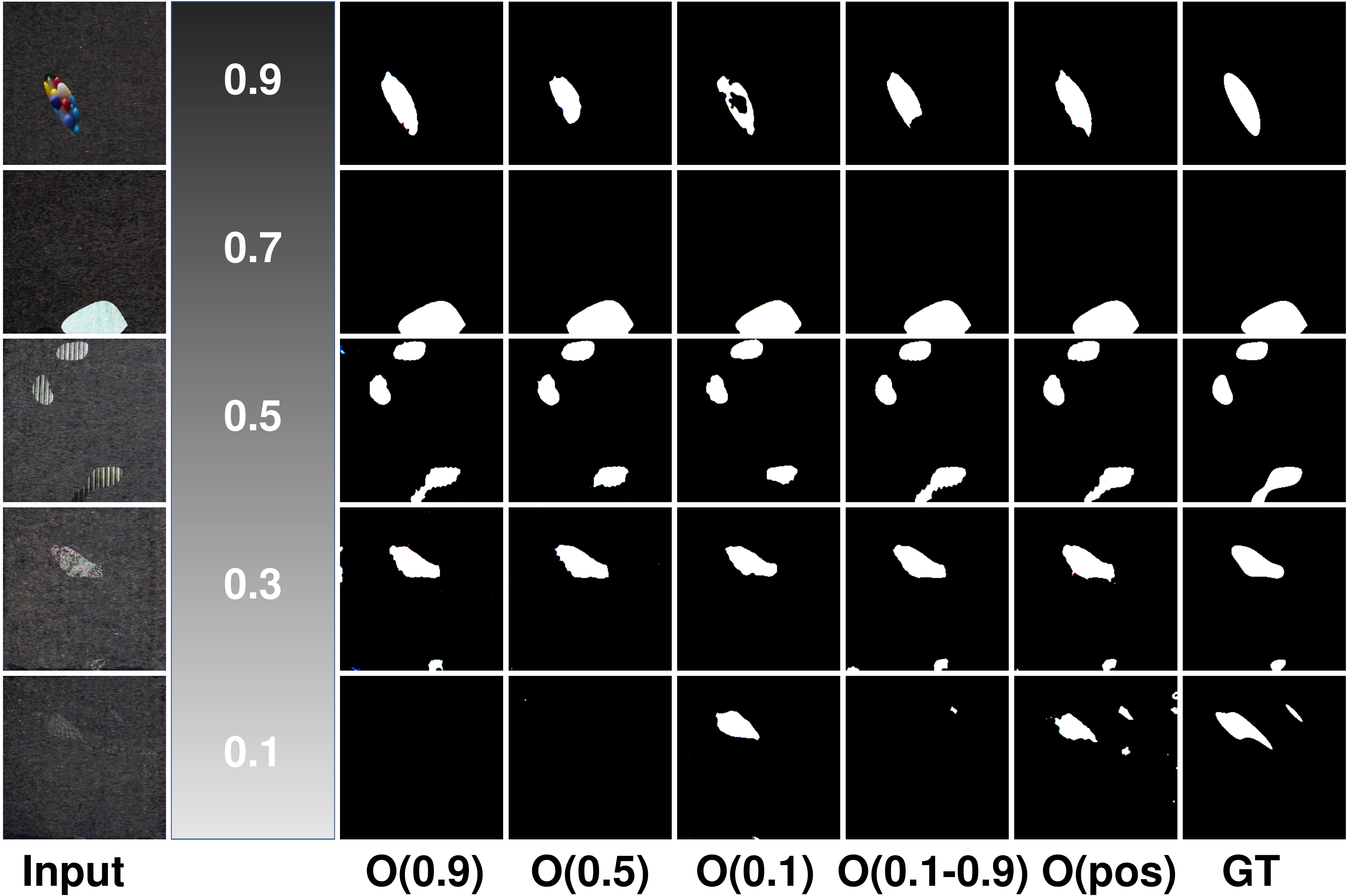}
		 \\[\smallskipamount]
		}
	  \caption{\label{fig: aug} Comparisons on the augmentation technique on dataset\cite{Bozic2021COMIND}. 
The simulated defects and a bar indicating their opacity are shown in the first and second columns.
The rest are the localization results from models trained with different opacity settings and the ground truth.
}
\end{figure}
In \cref{tab: ablation}, we present the results obtained by using different opacity settings.
Both \(G.S.C.M(0.1)\) and \(G.S.C.M(0.9)\) deliver unsatisfying results.
In experimental observations, obvious defects enable the network to learn defect localization faster and more stably.
The presence of more transparent defects, as discussed in \cref{sec: data}, encourages the model to discern subtle defects.
However, the training can be unstable because the generator struggles to identify transparent simulated defects in the initial stage.
Therefore, a progressive opacity strategy (Pos) is proposed in \cref{sec: data}.
It facilitates early training by simulating obvious defects and then gradually increasing the transparency of synthetic defects to promote the recognition of subtle defects.
Experimental results have demonstrated its effectiveness, with a performance of \textbf{68.9\%}, significantly outperforming fixed opacity values and random opacity settings.

The impact of opacity is further investigated by experimenting on simulated test sources with opacities ranging from 0.1 to 0.9.
In \cref{fig: aug}, models trained with high opacity cannot localize nearly transparent defects.
The models trained with low opacity perform adequately in the presented samples, but \(G.S.C.M(0.1)\) achieves only half the performance of \(G.S.C.M(0.9)\) in \cref{tab: ablation}, and it exhibits inconsistent performance in other samples.
Models trained with \(Pos\) demonstrate consistent performance regarding simulated test sources with different opacities, while also gaining optimal performance in \cref{tab: ablation}.
The results also suggest that performance could be improved by training with data of similar characteristics.
Since the model trains with unpaired samples, the model could directly utilize unlabeled real-world defects, which could be of less value in other methods. 
The accessibility to real-world defects could potentially lead to improved convergence of the model.


\section{Conclusion}
In this paper, we pointed out that over-reliance on nominal data may hinder the applicability of defect localization methods.
Our hypothesis posits that by removing explicit guidance during the training process, the model can develop the ability to discern defects in texture patterns under changing environments rather than solely acquiring superficial models that are heavily dependent on particular appearances learned during training.
We proposed a novel GAN-based framework that incorporates various constraints and implicit guidance to effectively localize defects.
The trained model successfully identifies defects without relying on modeling nominal sample subspaces.
The proposed method attains state-of-the-art performance on the benchmark datasets, while additional experiments substantiate its superior generalizability.
The potential of our framework can be further enhanced by leveraging raw defect samples from real-world scenarios to improve its performance, a practice uncommon in other methods that typically rely on defective samples accompanied by ground truth information during training.

{
    \small
    \bibliographystyle{ieeenat_fullname}
    \bibliography{main}
}


\end{document}